\documentclass{ceurart}

\usepackage{pifont}          % \ding{55} in tables
\usepackage[section]{placeins} % \FloatBarrier

\newcolumntype{L}[1]{>{\raggedright\arraybackslash\hspace{0pt}}p{#1}}

\newcommand{\cls}[1]{\textsc{#1}}
\newcommand{\edg}[1]{\texttt{#1}}
\newcommand{\state}[1]{\textsc{#1}}
\newcommand{\prop}[1]{\texttt{#1}}

\begin{document}

%% Rights management / conference information (CEUR-ART front matter).
\copyrightyear{2026}
\copyrightclause{Copyright for this paper by its authors.
  Use permitted under Creative Commons License Attribution 4.0 International (CC BY 4.0).}
\conference{KM4LAW 2026: 5th International Workshop on Knowledge Management
  and Process Mining for Law, 2026.}

\title{The Violation Situation Pattern: Persistent Representation of Compliance Violations in Knowledge Graphs}
% Venue note. OFF for the KM4LAW submission (the reviewers are KM4LAW; telling
% them it is under review at KM4LAW adds nothing). Uncomment BOTH lines for the
% arXiv v2 / preprint build, adjusting the wording to match its status.
%\tnotemark[1]
%\tnotetext[1]{Preprint. Under review at KM4LAW 2026.}

% ceurart's \parsename splits on spaces and takes the LAST token as the surname,
% which renders "Nima Kamali *Lassem*" and "(N. K. Lassem)". Bracing keeps the
% two-word surname together: "Nima *Kamali Lassem*" / "(N. Kamali Lassem)".
\author[1]{Nima {Kamali Lassem}}[%
  orcid=0009-0008-4326-9639,
  email=nima.kamali@dilitrust.com,
]
\cormark[1]
\author[1]{Fuqi Song}[%
  orcid=0009-0001-4900-5369,
  email=fuqi.song@dilitrust.com,
]
\author[2]{Seyid Amjad Ali}[%
  orcid=0000-0001-9250-9020,
  email=syedali@bilkent.edu.tr,
]
\address[1]{DiliTrust, Paris, France}
\address[2]{Department of Information Systems and Technologies,
  Bilkent University, T\"urkiye}
\cortext[1]{Corresponding author.}

\begin{abstract}
Existing compliance pipelines identify violations as transient query results, leaving no
persistent representation of the violation itself or its lifecycle, evidence, and audit
history. We address this limitation with the Violation Situation Pattern (VSP), a
reusable ontology pattern that extends the Situation pattern of Gangemi and Mika by
modeling each detected violation as a persistent first-class graph entity. Each
violation is associated with a rule identifier, temporal validity interval, lifecycle
state, and evidence links to the affected entities, while immutable lifecycle events
provide a complete and queryable audit trail. We instantiate VSP on the schema of a
deployed legal knowledge graph integrating corporate governance and contract data,
populated with public contract corpora containing violations with established ground
truth, and operationalize four deontic compliance rules. Rather than evaluating
detection algorithms, we evaluate the pattern's robustness under rule evolution. Results
show that repeated rule execution creates no duplicate violations, modifications to rule
definitions preserve complete audit histories, and expanding rule scope maintains
violation identity and lifecycle continuity. A case study using 73 regulatory
enforcement decisions further demonstrates that compliance rules can evolve to capture
additional confirmed violations without compromising the integrity of stored identities,
evidence, or audit histories. These findings show that VSP enables compliance knowledge
graphs to evolve without sacrificing traceability, explainability, or historical
consistency. The complete implementation, queries, SHACL shapes, and evaluation dataset
are publicly available to support the reproducibility of the reported results.
\end{abstract}

\begin{keywords}
  ontology design patterns \sep
  knowledge graph \sep
  legal compliance \sep
  violation lifecycle \sep
  provenance \sep
  audit \sep
  deontic logic \sep
  SHACL
\end{keywords}

\maketitle

\section{Introduction}
\label{sec:intro}

Compliance pipelines typically execute detection queries that return a set of flagged
results without preserving the violations themselves as persistent graph entities. Once
a query completes, the detected violations disappear, leaving information such as
detection time, review status, and resolution history to be managed outside the
knowledge graph. When the same query is executed again, previously identified violations
are simply re-derived rather than recognized as existing ones. As a result, the system
cannot distinguish a recurring violation from a newly detected one. Although lifecycle
information can be maintained in an external repository, doing so fragments the audit
trail and requires every compliance investigation to integrate information across
multiple systems.

This limitation becomes apparent in a simple compliance scenario. A signature-authority
breach is detected for a contract in January and subsequently reviewed and dismissed.
When the rule is executed again in April, the same breach is detected once more. An
auditor later asks when the issue was first identified, who dismissed it, and whether it
has recurred since. Detection outputs alone cannot answer these questions because the
January detection, the review decision, and the April re-detection were never captured
as the lifecycle of a single persistent violation.

Throughout this paper, compliance rules are expressed in a deontic contract logic, the
knowledge graph is implemented as a property graph, and portability is assessed using
the Shapes Constraint Language (SHACL). Existing compliance approaches, including
SHACL-based validation~\cite{david2025shacl} and deontic reasoning over reified General
Data Protection Regulation (GDPR) representations~\cite{robaldo2020dapreco}, effectively
detect violations when the required information is available. However, they treat
violations as transient query results rather than durable graph entities. By contrast,
knowledge graphs represent domain concepts as persistent semantic
objects~\cite{hogan2021kg}, and legal knowledge graphs extend this principle to
contracts, clauses, persons, and
signatures~\cite{filtz2021linkedlegal,leone2020stocklegal}. Yet the violations detected
over these entities remain ephemeral: a rule match identifies a violation, but the graph
records neither when it was detected, who reviewed it, nor how it was ultimately
resolved. Consequently, repeated detections cannot be linked to a common identity,
making long-term auditing and lifecycle management difficult.

To address this limitation, we introduce the Violation Situation Pattern (VSP), which
represents every detected violation as a persistent graph node. Each violation is
associated with a rule identifier, a temporal validity interval governed by a staleness
constraint, a finite-state lifecycle for review and remediation, and evidence relations
linking the violation to the relevant contracts and persons. VSP builds on the Situation
Ontology Design Pattern (ODP) proposed by Aldo Gangemi and Peter
Mika~\cite{gangemi2003situations} while extending it with lifecycle management,
staleness handling, and explicit identity conditions required for compliance auditing.
Every lifecycle transition is recorded as an immutable graph event, allowing the
complete history of a violation to be reconstructed through graph traversal.

Representing violations as persistent graph entities directly expands the questions
that a compliance knowledge graph can answer. Repeated detections are linked to an
existing violation rather than creating duplicate records, review decisions become part
of the graph together with the responsible actor, rationale, and timestamp, and audit
queries can be answered directly within the same graph that stores contracts, persons,
and other domain entities. Building on these capabilities, this paper makes the
following contributions:

\begin{description}
\setlength{\itemsep}{2pt}
\item[C1.] We introduce the Violation Situation Pattern (VSP), a reusable ontology pattern that
represents each compliance violation as a persistent graph node with a stable identity,
temporal validity interval, lifecycle state, and evidence relations to the entities
referenced by the governing rule.

\item[C2.] We define a lifecycle model in which every state transition is recorded as an immutable
event containing the responsible actor, rationale, and timestamp, enabling complete
review histories to be reconstructed through graph traversal.

\item[C3.] We empirically evaluate the pattern's stability on a knowledge graph containing
violations with known ground truth, demonstrating that repeated rule execution creates
no duplicate violations, rule revisions preserve transition histories, and widening a
rule's scope maintains violation identity and lifecycle continuity.

\item[C4.] We present a worked case in which a compliance rule is extended to capture additional
confirmed violations, while previously recorded violations retain their identity,
evidence, and complete lifecycle history. The extended rule is evaluated against 73
regulatory enforcement decisions.

\item[C5.] We demonstrate the portability of the pattern through a cross-formalism verification
using SHACL, confirming that VSP is not tied to a property-graph implementation.

\end{description}

Together, these contributions provide the properties required for an auditable
compliance representation that remains stable as compliance rules evolve over time. The
remainder of the paper is organized as follows. Section~\ref{sec:background} defines the problem in legal
compliance systems and identifies the properties required for compliance-grade
representations. Section~\ref{sec:related} reviews related work on compliance management, norm
formalization, situation modeling, and provenance, showing that although each addresses
part of the problem, none provides the complete set of properties needed for persistent
compliance auditing. Section~\ref{sec:vsp} introduces the Violation Situation Pattern, Section~\ref{sec:rules}
operationalizes four compliance rules, Section~\ref{sec:eval} evaluates the pattern and presents a
worked case study, Section~\ref{sec:limits} discusses limitations, threats to
validity, and future work, and Section~\ref{sec:conclusion} concludes the paper.
\section{Background and Problem Setting}
\label{sec:background}

\label{sec:bg-legaltech}
Compliance is managed as an organizational process spanning strategy,
processes, technology, and people \cite{racz2010grc}, and research on business
process compliance has concentrated on aligning regulatory rules with the
design, verification, and validation of processes \cite{hashmi2018bpc}. In
operational legal-tech systems this takes the form of an alert-driven
workflow: rules run as queries over contracts, governance data, or process
records, and the resulting alerts pass to compliance personnel for review,
escalation, dismissal, and remediation. Compliance practice separates these
functions at the tool level. The compliance management ontology of Abdullah et
al.~\cite{abdullah2016cmo}, built from practitioner interviews and eight
organizational case studies with heads of compliance and compliance managers,
models review and record keeping as program activities of their own and
distinguishes review tools and record-keeping and reporting tools from
obligations-management tools. In the deployments this work draws on, the
consequence is that a finding's review state is held in ticketing or
case-management systems while the contracts, mandates, and persons it concerns
are held in the graph. During regulatory review, however, an
organization must demonstrate when a violation was first detected, who
assessed it, and how it was resolved, and that history is scattered across
emails, spreadsheets, and case-management tools.

\label{sec:bg-properties}
A compliance-grade representation requires three properties. \textbf{Stable identity}
ensures that re-evaluation recognizes an existing violation rather than
creating a duplicate. \textbf{Lifecycle management} records the current review stage.
\textbf{Immutable provenance} makes every decision part of a reconstructible audit
history. None is difficult in isolation: validation reports record findings,
alert repositories assign identifiers, and ticketing systems track review
state. The difficulty is that none of them holds the entities the rule reasons
over, so an audit question means correlating a ticket, a report, and a graph, and
keeping that correspondence accurate as rules change. What a compliance-grade
representation adds is co-location: the violation record lives in the same graph
as the contracts, clauses, and persons it concerns.

\label{sec:bg-lemclm}
DiliTrust's Legal Entity Management (LEM)\footnote{\url{https://www.dilitrust.com/legal-entity-management-roi-calculator/}} captures corporate governance facts: companies, governance bodies, mandates with appointment and cessation dates, and signing authorities. Contract Lifecycle Management (CLM)\footnote{\url{https://www.dilitrust.com/dilitrust-governance-suite/contract-management/}} captures contract-level facts, including contract metadata, contracting parties, signatures, and clauses by typology.
Clause-boundary and typology extraction that feeds CLM ingestion is inspired by contract NLP benchmarks such as
CUAD~\cite{hendrycks2021cuad} and
ContractNLI~\cite{koreeda2021contractnli}. Graphs of this shape
support entity-level queries (who serves on which board, which
contracts a company has signed), but compliance checks are queries
over those entities rather than facts within the graph. A query that
returns a contract signed without an active mandate surfaces a fact;
that fact does not enter the data as a persistent object.

\section{Related Work}
\label{sec:related}

% ---------------------------------------------------------------------
% P1 — Compliance management and GRC (the operational context).
% ---------------------------------------------------------------------
Compliance management has been studied as an organizational
discipline before it was studied as a data-modeling problem. The governance
frame of reference of Racz et al.~\cite{racz2010grc} and the business-process
compliance survey of Hashmi et al.~\cite{hashmi2018bpc}, both introduced in
Section~\ref{sec:bg-legaltech}, establish what has to happen once a violation is
found and supply the vocabulary for describing it; the compliance management
ontology of Abdullah et al.~\cite{abdullah2016cmo} defines a corrective control
as remediation of a non-compliance that has been detected, yet the detected
non-compliance is not itself among its constructs. What this literature does not
supply is a representation of the violation itself as a persistent, queryable
object: the review state, the detection history, and the evidence linking a
finding to the entities it concerns remain artifacts of the surrounding tooling
rather than facts in the data.

% ---------------------------------------------------------------------
% P2 — Formalization and detection. Justified first introduction of FCL.
% ---------------------------------------------------------------------
A second line of work formalizes the norms themselves so that non-compliance
can be derived. Formal Contract Logic (FCL), introduced by Governatori and
Milosevic~\cite{governatori2006fcl}, represents contractual obligations in a
defeasible deontic setting and accounts for the secondary obligations that
arise once a primary obligation is breached; the same formalism has been
applied to checking business processes against contract
constraints~\cite{governatori2006compliance}. We adopt FCL for the rule layer
of this work for that reason: it expresses the obligation structure of our
rules directly, and it keeps the rule body separable from whatever stores the
result. Robaldo et al.~\cite{robaldo2020dapreco} formalize GDPR provisions in
reified input/output logic and publish them as the DAPRECO knowledge base,
where a violation is a logical consequence of a failed obligation rather than a
stored entity. Comparable observations hold for validation-based approaches:
SHACL-driven contract compliance checking~\cite{david2025shacl} reports
violations through a validation report that is regenerated on each run. Across
this line of work the output of detection is a determination, and the
determination does not outlive the query that produced it.

% ---------------------------------------------------------------------
% P3 — Situation modeling. RSS and the Description half.
% ---------------------------------------------------------------------
Within ontology engineering, the reification of contexts is well established.
Gangemi and Mika~\cite{gangemi2003situations} introduce Descriptions and Situations, a
two-level architecture in which a Description is a non-physical entity such as
a norm or a plan, and a Situation is a reified state of affairs that satisfies
it through structured constituents. VSP inherits this architecture directly.
The deontic rule plays the role of the Description, and each detected violation
is the Situation that satisfies it. We do not reify the Description inside the
graph: keeping the rule external, referenced only
by identifier, is what allows a rule body to be revised without disturbing the
violations already recorded under it. Carriero et
al.~\cite{carriero2021recurrent} define the RecurrentSituationSeries pattern
for situations that recur at regular intervals and share invariant unifying
factors. Repeated detections of the same violation are recurrence in exactly
this sense, and could be modeled as a series whose members are the individual
detection events. We take the opposite decision deliberately: rather than
distributing a finding across a series, VSP unifies its detections under one
identity and records their succession in the transition log, because an auditor
asks about the finding rather than the series. The two treatments differ in
granularity rather than in subject matter, and RecurrentSituationSeries supplies
neither the identity criterion that merges repeated detections into one object
nor the lifecycle over which that object is reviewed and closed.

% ---------------------------------------------------------------------
% P4 — Audit and provenance graphs.
% ---------------------------------------------------------------------
Closer to the audit motivation of this work, Waltersdorfer and
Sabou~\cite{waltersdorfer2025auditing} argue that knowledge graphs can address
the fragmentation of audit evidence in AI system governance, and they set out
architectural directions for collecting and integrating audit data
semantically. Their contribution is programmatic rather than schematic: the
individual audit finding is not given a graph structure, an identity condition,
or a state. At the infrastructure level, Dibowski~\cite{dibowski2024prov}
presents PROV-STAR, an RDF-star extension of PROV-O that intercepts updates and
records every inserted and deleted triple with its timestamp and agent, so that
any prior state of a graph can be recovered. PROV-STAR tracks mutations of
triples in a domain-agnostic way; VSP records domain events, where each entry
in the log is a named lifecycle transition of a violation rather than a raw
edit. The two are compatible, and a PROV-STAR style substrate would be a
reasonable persistence layer for the transition log described in
Section~\ref{sec:vsp}.

% ---------------------------------------------------------------------
% P5 — Norm-layer graphs.
% ---------------------------------------------------------------------
Knowledge graphs have also been applied to the norms themselves. Hernandez et
al.~\cite{hernandez2025tair} introduce the TAIR ontology, which extracts
normative statements from the EU AI Act and from harmonized ISO standards and
publishes them as linked Requirement and Concept instances that can be queried
for cross-document consistency. TAIR operates on regulatory text and reports no
enforcement instances; whether a given organization has violated a requirement
is outside its scope. The relation to VSP is one of layering rather than
competition. A requirement represented in a norm graph of this kind is a
natural source for the rule identifier that a violation node carries, and the
two representations can be linked without either absorbing the other.
% MOVED here per item 3.2 (ARGOS sentence relocated from old Section 3 opening):
A parallel effort from the ontology-patterns workshop WOP 2025,
ARGOS~\cite{pathirage2025argos}, targets governance patterns for
LLM-driven applications and data operations; it governs the
operations that produce findings rather than the findings themselves.

% ---------------------------------------------------------------------
% P5b — Engineering practice. Answers the "this is just event sourcing" reading.
% ---------------------------------------------------------------------
One further body of work sits outside the research literature. Event sourcing ---
recording state as an append-only sequence of events and deriving the current
state by replaying them --- is established engineering practice, and VSP's
transition log is structurally an event log of that kind. The resemblance is
real, but the problems differ. In event sourcing an object's identity is fixed by
the command that creates it, so the log stores an asserted fact, and revising the
schema of those events is correspondingly expensive, requiring conversion or
replay of the store~\cite{overeem2017eventsourcing}. A violation is not asserted
but derived: it exists because a rule matched, and the rule is revisable. VSP
therefore keys identity on the referents a rule concerns rather than on the rule
that produced the match --- which is what lets a rule body be rewritten while the
findings recorded under it keep their identity and history, with no conversion
and no replay.

% ---------------------------------------------------------------------
% P6 — Synthesis / gap statement.
% ---------------------------------------------------------------------
Read together, these lines of work divide along a consistent boundary. The
norm-layer approaches represent obligations, the detection approaches decide
whether an obligation has been breached, the governance literature prescribes
what an organization then does, the provenance approaches record how a
graph changed, and engineering practice preserves asserted state. What none of the approaches surveyed here provides is a
representation of the finding that sits between them: an object with an identity stable across repeated
evaluations of the same rule, a lifecycle recording the review and remediation
decisions taken about it, an append-only history of those decisions, and
evidence relations to the entities the rule concerns. In an operational legal
graph it is this object that an auditor asks about, and this object that is absent.
VSP supplies it at the application level, and it does so without prescribing how
detection is implemented or where the graph is stored.

% ---------------------------------------------------------------------
% Table 1 handover. Demoted from "gap analysis" to design rationale.
% ---------------------------------------------------------------------
Before defining the pattern we make the reuse strategy explicit. Several
catalogue patterns supply part of what a violation record requires, and
Table~\ref{tab:odp-gap} summarizes which part each one supplies. The comparison is
offered as design rationale rather than as evidence of a gap in the catalogue:
it records why no single pattern could be adopted unchanged, and which
constituents of VSP are inherited rather than new.

\begin{table}[!htbp]
\caption{Reuse analysis of catalogue patterns. Columns: ID = stable identity
across re-evaluations; LC = finite-state lifecycle with named
transitions; AT = immutable audit/transition log; EV = multi-referent
evidence cardinality. \checkmark{} = directly provided;
$\sim$ = partially provided; \ding{55}{} = not provided.}
\label{tab:odp-gap}
\centering\footnotesize
\renewcommand{\arraystretch}{1.05}
\setlength{\tabcolsep}{5pt}
\begin{tabular}{L{5.6cm}cccc L{5.4cm}}
\toprule
\textbf{Pattern (source)} &
\textbf{ID} & \textbf{LC} & \textbf{AT} & \textbf{EV} &
\textbf{Constituent supplied / not supplied}\\
\midrule
Situation / Description+Situation~\cite{gangemi2003situations} &
$\sim$ & \ding{55} & \ding{55} & \checkmark &
No identity, lifecycle, or transition log \\
EventCore~\cite{krisnadhi2017eventcore} &
\ding{55} & \ding{55} & \ding{55} & $\sim$ &
Point-in-time only; no lifecycle or AT log \\
RecurrentSituationSeries~\cite{carriero2021recurrent} &
\ding{55} & \ding{55} & $\sim$ & $\sim$ &
Series-level recurrence structure; no per-instance identity or lifecycle \\
TimeIndexedSituation~\cite{presutti2008contentodp} &
$\sim$ & \ding{55} & \ding{55} & \checkmark &
Adds $T$; no $S$ or AT log \\
TimeIndexedPersonRole~\cite{presutti2008contentodp} &
\checkmark & \ding{55} & \ding{55} & $\sim$ &
Identity per role-period; no lifecycle \\
N-ary Relations~\cite{noy2006naryrelations} &
\checkmark & \ding{55} & \ding{55} & \checkmark &
Closest structural sibling; no LC or AT log \\
PROV-O \texttt{prov:Activity}~\cite{provo2013} &
$\sim$ & $\sim$ & \checkmark & $\sim$ &
Provenance trail; no native LC states or transition table per situation \\
\midrule
\textbf{VSP (this paper)} &
\checkmark & \checkmark & \checkmark & \checkmark &
Composite identity key, five-state automaton, append-only transition log,
rule-specific evidence edges \\
\bottomrule
\end{tabular}
\end{table}

\section{The Violation Situation Pattern}
\label{sec:vsp}

\subsection{Formal Definition}
\label{sec:vsp-formal}

Detection creates the violation record; review and remediation transition it;
auto-revalidation re-confirms it or, on re-detection after closure, reopens it.
We refer to the reified object as a violation record, and to the class that
realizes it in a property graph as \cls{ViolationInstance}.

We define the Violation Situation Pattern as the tuple
\begin{equation}
\mathrm{VSP} = (r, T, S, E)
\end{equation}
where $r$ is the FCL (Section~\ref{sec:related}) rule identifier the
violation instantiates, $T = [\prop{detectedAt},\allowbreak\,\prop{validUntil}]$ is the
temporal validity interval, $S \in \{\state{DETECTED},$ $\state{CONFIRMED},$
$\state{UNDER\_REVIEW},$ $\state{REMEDIATED},$ $\state{DISMISSED}\}$
is the lifecycle state, and $E$ is a set of evidence edges binding
the violation to its referent entities.

Following the ODP catalogue style of Presutti and
Gangemi~\cite{presutti2008contentodp}:

\smallskip
\noindent\textbf{Context.} A knowledge graph in which a recurring
rule-driven check produces detection events that consumers need to
query historically without re-executing the rule. Nothing in that description is
specific to legal compliance; it is what a domain must satisfy to instantiate the
pattern.

\noindent\textbf{Solution.} Reify each detection as a
\cls{ViolationInstance} node carrying $(r, T, S, E)$. Pair it with an
immutable \cls{LifecycleTransition} class so the projection (current
\cls{ViolationInstance} state) and the event log
(\cls{LifecycleTransition} history) are both graph objects. Enforce
two invariants at the application layer: only transitions in the
lifecycle automaton are accepted, and \prop{detectedAt} is immutable.

\noindent\textbf{Consequences.} The realization carries two tradeoffs. Both
invariants are enforced at the application layer rather than the schema level, so
a writer that bypasses the mediator of Section~\ref{sec:vsp-graph} can violate
either. And a property-graph realization does not provide Web Ontology Language (OWL) entailment,
so class and property inferences available in an RDF stack are not available
here. Section~\ref{sec:eval-shacl} shows that the pattern nonetheless transfers to an RDF-star
and SHACL realization, which recovers schema-level enforcement.

\smallskip
\noindent\textbf{Competency questions.} The composition of $(r, T, S, E)$ supports audit
questions that stateless detection cannot answer in a single traversal. The
questions below were derived from requirements communicated by DiliTrust's legal
and product teams, and confirmed by them as representative of what is asked in
practice; they correspond to the accountability demands set out in
Section~\ref{sec:bg-legaltech}, where an organization must show when a violation was
detected, who assessed it, and how it was resolved. Four are representative:

\begin{description}
\setlength{\itemsep}{2pt}
\item[CQ1.] Was a given violation previously dismissed and later re-detected? A path
query over the transition log for a \state{DISMISSED}-to-\state{DETECTED} sequence.
\item[CQ2.] Which violations were active at a given time $t$? A window query over
violations with $\prop{detectedAt} \leq t \leq \prop{validUntil}$.
\item[CQ3.] Which entities are jointly implicated in two or more open violations? A
multi-hop join across evidence edges with a threshold on the count.
\item[CQ4.] What is the complete review history of a given violation? An ordered
retrieval of its transition log, returning actor, timestamp, and reason.
\end{description}

Section~\ref{sec:eval-stability} reports their execution against the evaluation graph.

\smallskip
\noindent\textbf{Properties under re-evaluation.} Two properties follow from the identity key
and the append-only log. First, re-evaluation is idempotent: because a violation
is written through an upsert on its composite identity key (Section~\ref{sec:vsp-graph}), a
repeated detection of the same violation matches the existing node and extends
its validity interval rather than creating a second node. Second, audit history
is reconstructible: because the transition log is append-only (Section~\ref{sec:vsp-graph}), the
complete sequence of states a violation has occupied is recoverable by a single
traversal of its transition nodes ordered by timestamp, regardless of how many
re-evaluations have occurred. No single catalogue pattern in Table~\ref{tab:odp-gap}
supplies all four of stable identity across re-evaluations, a finite-state
lifecycle, an immutable transition log, and multi-referent evidence. Specifying
them together on one situation node, with the rule left external and therefore
revisable, is the pattern's contribution.

\subsection{Graph Realization}
\label{sec:vsp-graph}

\begin{figure}[!htbp]
\centering
\includegraphics[width=0.60\textwidth]{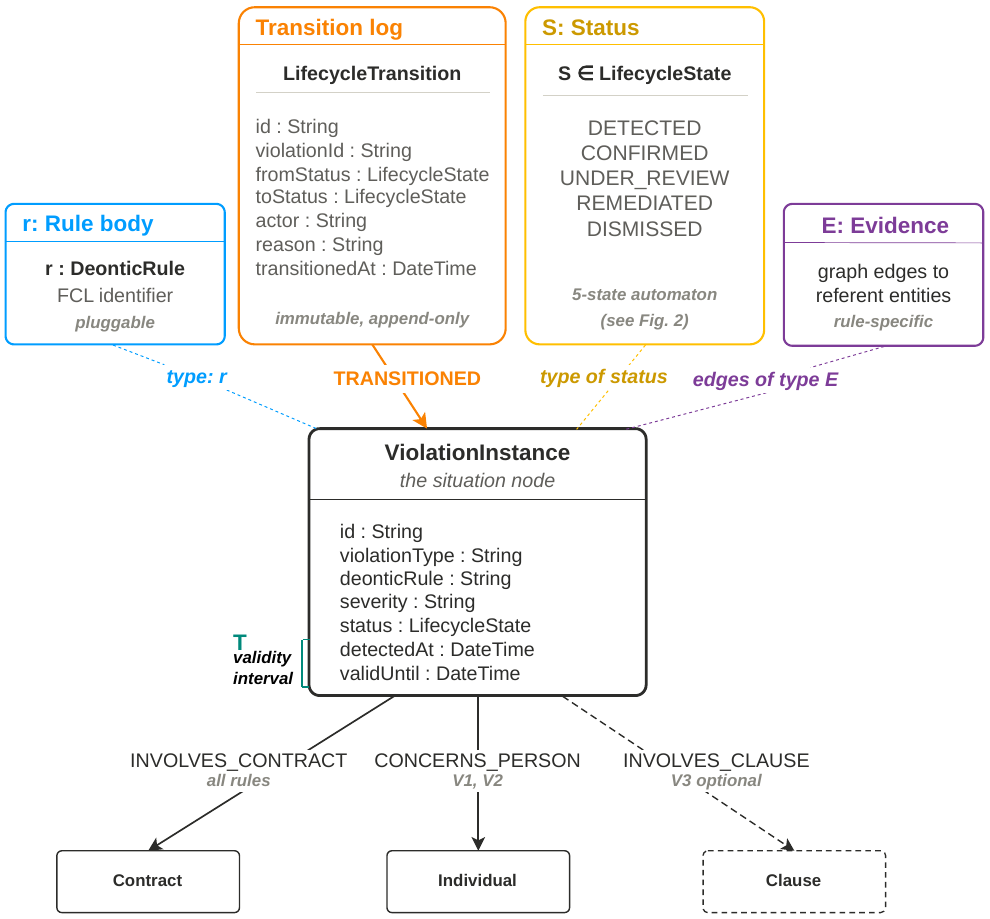}
\caption{Pattern components $(r, T, S, E)$ around the
\cls{ViolationInstance} node. Evidence cardinality is rule-specific; per-rule
details are in Section~\ref{sec:v3v4}.}
\label{fig:vsp-a}
\end{figure}

The tuple defines the pattern abstractly; the property-graph
realization below is one instantiation, and Section~\ref{sec:eval-shacl}
shows the same pattern carries to RDF stacks via a SHACL+PROV-O
retrofit. As Fig.~\ref{fig:vsp-a} shows, VSP realizes in a property graph as two node classes ---
\cls{ViolationInstance}, the situation node, and
\cls{LifecycleTransition}, the immutable event-log node --- linked by a
\edg{TRANSITIONED} edge, together with domain-specific evidence edges (in the
legal domain, \edg{INVOLVES\_CONTRACT}, \edg{CONCERNS\_PERSON},
\edg{INVOLVES\_CLAUSE}).

The node carries \prop{deonticRule}, the identifier $r$ of the rule whose match
produced it, alongside a \prop{violationType} category and a \prop{severity} label.
Identity is composite: $(\mathrm{violationType}, \mathrm{contract\_id},
\mathrm{person\_id})$ for V1/V2 and $(\mathrm{violationType},
\mathrm{contract\_id})$ for V3/V4 --- keyed on the category and its referents, not
on \prop{deonticRule}, which is what leaves the rule body free to change. Detection writes each violation through
an upsert keyed on this identity: a write that creates the node if no node
carries that key and updates the existing one otherwise. On the first match the
node is created with its detection timestamp; on any later match the same node is
retained, its validity interval is extended, and its detection timestamp is never
modified. This is the mechanism behind the idempotence property of Section~\ref{sec:vsp-formal}.

Two invariants govern every subsequent write. Only arcs of the lifecycle
automaton (Section~\ref{sec:vsp-lifecycle}) are committed, and \prop{detectedAt} is
immutable across an instance's lifetime. All state changes pass through a single
write path, the \texttt{transition()} mediator, which accepts a change only if it is
a permitted arc and never alters \prop{detectedAt}. Because a property graph does not
enforce these conditions at the schema level, the mediator enforces them in the
application layer, and a caller bypassing it with a direct write can violate either
invariant or both; Section~\ref{sec:limits-system} states the consequence, and
schema-level enforcement is achievable on RDF-star\,+\,SHACL
stacks~\cite{shacl2017,rdfstar2024}.

\subsection{Lifecycle Automaton}
\label{sec:vsp-lifecycle}

\begin{figure}[!htbp]
\centering
\includegraphics[width=0.46\textwidth]{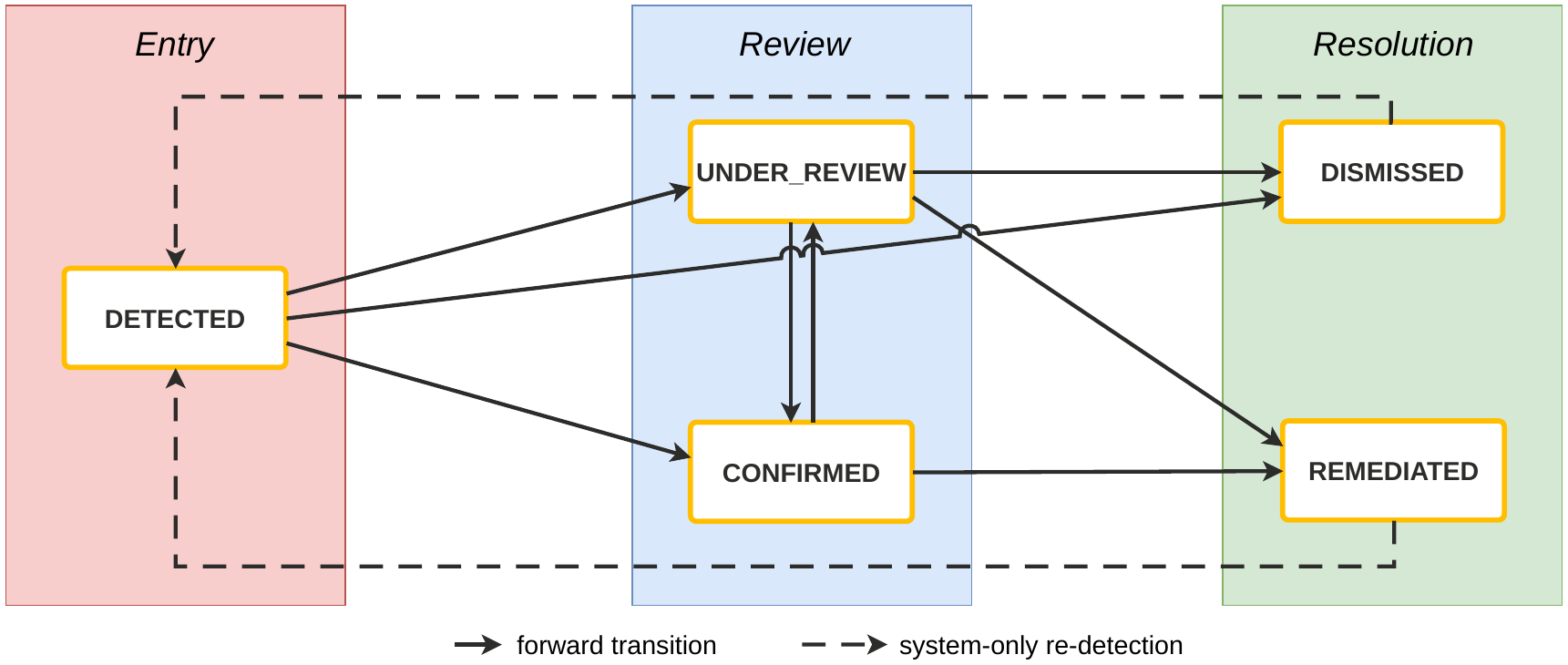}
\caption{The five-state lifecycle automaton. Dashed arrows are the two
system-only re-detection arcs, from \state{REMEDIATED} and \state{DISMISSED} back
to \state{DETECTED}.}
\label{fig:vsp-b}
\end{figure}

The state machine has five states and ten arcs, shown in Fig.~\ref{fig:vsp-b}: from
\state{DETECTED} to \state{UNDER\_REVIEW}, \state{CONFIRMED}, or \state{DISMISSED};
from \state{UNDER\_REVIEW} to \state{CONFIRMED}, \state{REMEDIATED}, or
\state{DISMISSED}; from \state{CONFIRMED} to \state{UNDER\_REVIEW} or
\state{REMEDIATED}; and from \state{DISMISSED} or \state{REMEDIATED} back to
\state{DETECTED}.
Actor type is recorded on each \cls{LifecycleTransition} node rather than
constrained per arc: arcs into \state{UNDER\_REVIEW} and \state{DISMISSED} are
taken by a human reviewer, the rest by either reviewer or revalidator. Each
accepted transition writes a \cls{LifecycleTransition} node with a
\edg{TRANSITIONED} edge to the \cls{ViolationInstance}. The revalidation cycle re-executes each rule on a
schedule: if the underlying condition no longer holds, the violation's validity
interval is left to lapse, and if a closed violation's condition holds again, the
re-detection arc reopens it.

Re-detection raises an identity question: is a violation that was dismissed
and later re-detected the same violation? Under the identity criterion it is.
The composite key matches, so the record is reopened rather than duplicated,
and the earlier dismissal, with its stated reason, remains in the transition
log. What begins is a new review episode of the same violation, and it is
precisely this construction that makes recurrence auditable: the graph can
show that this finding was judged not actionable once before, by whom, and
why. The arc from \state{UNDER\_REVIEW} directly to \state{REMEDIATED} exists for
the same reason: requiring confirmation first would make it impossible to record
a correction that arrived before the review concluded.

\section{Rule Operationalization}
\label{sec:rules}

Each rule is translated into a graph query and written through the
identity-preserving upsert of Section~\ref{sec:vsp-graph}, which instantiates a
persistent violation record. V1 and V2 are expressed as FCL obligations, V3 and V4
through the clause-presence operationalization below. Table~\ref{tab:rules} lists the rules.
The four were selected to exercise different parts of the pattern rather
than to cover a legal area exhaustively, varying along three axes:

\begin{description}
\setlength{\itemsep}{1pt}
\item[Subgraph and evidence cardinality.] V1 and V2 operate on the governance
subgraph and attach two evidence referents, a contract and an individual; V3 and
V4 operate on the contract subgraph, V3 attaching an optional clause referent
and V4 a single contract referent.
\item[Normative source.] Statutory obligations for V1, V2, and V4; a
policy-derived contractual control for V3.
\item[Rule form.] The full FCL expression for V1 and V2; the clause-presence
operationalization for V3 and V4, used because the data layer represents clauses
by \prop{clauseType} rather than by structured obligation flags. We discuss this
rule-correctness boundary in Section~\ref{sec:v3v4}.
\end{description}

\noindent V4 carries the external evaluation because it is the only rule for which
regulator-labeled ground truth is publicly available.

\begin{table}[!htbp]
\caption{Four implemented rules. V1, V2, V4 are statutory; V3 is
policy-derived (see Section~\ref{sec:v3v4}).
Statutory rules derive from binding legal instruments; the policy-derived rule
expresses an internal contractual control. The distinction is recorded because
the pattern treats both identically: identity, lifecycle, and evidence semantics
do not depend on the normative force of the rule.}
\label{tab:rules}
\centering\footnotesize
\renewcommand{\arraystretch}{1.05}
\setlength{\tabcolsep}{5pt}
\begin{tabular}{L{5.3cm}lL{7.2cm}}
\toprule
\textbf{Rule} & \textbf{Type} & \textbf{Source}\\
\midrule
V1 Unauthorized Signature & statutory &
Ultra vires; Dir.\ 2009/101/EC Art.\,9; CdC L225-35 \\
V2 Expired Signatory Mandate & statutory &
CdC L225-35 (corporate authority expiry) \\
V3 Missing Confidentiality Clause & policy &
Contractual control (commercial practice) \\
V4 Missing DPA Breach-Notification & statutory &
GDPR 28(3)(f), 33; SCC 2021/915~\cite{ec2021scc} Art.\,9 \\
\bottomrule
\end{tabular}\\[2pt]
{\footnotesize CdC: French Code de Commerce. DPA: Data Processing Agreement. SCC: Standard Contractual Clauses.}
\end{table}

\subsection{V1 and V2: Signature-Authority Violations}

V1 (unauthorized signature) obliges that every signatory hold a mandate at
the contracting party whose scope includes \textsc{Signature\_Authority} and whose
temporal validity covers the signing date; a violation is the negation of
that consequent for any non-\textsc{Other} signatory role. Detection identifies
signatures by individuals whose \edg{AUTHORIZED\_BY} mandates either lack the
required scope or are out of temporal validity at signing. V2 (expired signatory
mandate) differs in that the signer once held the required mandate: detection
finds an \edg{AUTHORIZED\_BY} mandate whose end date precedes the signing date and
confirms no replacement mandate covers it. The write step is structurally
identical, differing only in \prop{violationType}. The legal distinction is
substantive, ultra vires from inception versus action beyond expiry, and the two
are kept as separate rules for that reason.

\subsection{V3 (Missing Confidentiality) and V4 (Missing DPA Breach-Notification)}
\label{sec:v3v4}

V3 flags contracts lacking a \textsc{Confidentiality} clause. The evaluation
set comprises commercial contracts and contains no non-disclosure or
data-processing agreements, for which the check is not meaningful.
V3 is policy-derived: confidentiality
is a standard contractual control rather than a statutory obligation
and is labeled a violation for uniformity with V1/V2/V4 rather than
normative parity. V3 demonstrates that VSP applies identically to
policy-derived rule bodies: the pattern's identity, lifecycle, and
evidence semantics are independent of the rule's legal status. V4
flags DPAs lacking a \textsc{Breach\_Notification} clause per SCC
2021/915 Art.\,9. V4 detects clause presence rather than clause
compliance. A DPA with a 90-day notification window passes V4 but
violates GDPR Art.\,33's 72-hour requirement. This is a
rule-correctness boundary, not a pattern boundary;
Section~\ref{sec:eval-gdprhub} quantifies the share of
regulator-issued findings this operationalization covers.

Evidence cardinality is rule-specific: V1 and V2 attach \edg{INVOLVES\_CONTRACT}
and \edg{CONCERNS\_PERSON} edges to \cls{Contract} and
\cls{Individual} nodes respectively; V3 attaches
\edg{INVOLVES\_CONTRACT} and, when the contract carries clause nodes,
an \edg{INVOLVES\_CLAUSE} edge to a \cls{Clause} node; V4 attaches
\edg{INVOLVES\_CONTRACT} only.

\section{Evaluation}
\label{sec:eval}

The graph, the detection queries, the SHACL shapes, the evaluation datasets, and
the scripts that produce every result in this section are available as a public
artifact.\footnote{\url{https://doi.org/10.5281/zenodo.21769799}} Each result below is reproduced by running the corresponding
script against the released graph.

The evaluation supports the pattern claim through three pieces of evidence of
distinct types: a direct check of the pattern's stability properties, a rule-body
scope coverage analysis for V4 against GDPRhub decisions, and a cross-formalism
portability check on V3/V4.

The evaluation graph instantiates the LEM/CLM schema of
Section~\ref{sec:bg-lemclm} but is populated with contracts from public corpora ---
CUAD~\cite{hendrycks2021cuad}, ContractNLI~\cite{koreeda2021contractnli}, and the
EU Standard Contractual Clauses~\cite{ec2021scc} --- rather than with client data.
Those corpora carry no corporate-governance records, so the companies, individuals,
mandates, and signing authorities are generated, and violations are planted so that
ground truth is known for every case. It carries 291
\cls{Company}, 269 \cls{Individual}, 198 \cls{Mandate}, 235
\cls{Contract}, 331 \cls{ContractParty}, 100 \cls{Clause}, and 140
\cls{ViolationInstance} nodes, plus 19 \cls{LifecycleTransition} nodes
from a deterministic lifecycle exercise reaching all five states.
V1, V2, and V3 are scoped through a manifest that partitions contracts into an
authorized group, a planted-violation group, and an expired-mandate group; the
partition is what makes ground truth available, so the detection counts below are
counts over that partition. The stability checks of
Section~\ref{sec:eval-stability} do not depend on it, and are reported unscoped
where the query planner permits.

\subsection{Pattern Stability}
\label{sec:eval-stability}

The pattern's central claims are that re-evaluation does not duplicate a
violation, that the audit log survives a change to the rule body, and that a
violation keeps its identity and history when a rule's scope is widened.

The first two follow from the construction: an upsert on a composite key cannot
create a second node for a key that already exists, and a log that is only ever
appended to cannot lose an entry. What is not guaranteed is that the
implementation realizes them, and that they survive operations which could
plausibly break them. Each check below runs against the canonical graph, restores
it on completion, and is reproducible from the artifact. Idempotence is measured
by re-running every detection rule and comparing the violation set before and
after; log immutability by hashing every transition node across both the
re-evaluation and the rule revision.

Re-evaluation left the graph at 140 violations with no node created or lost
and no \prop{detectedAt} altered; all 19 transition-log entries survived both the
re-evaluation and the rule revision unchanged. Because idempotence is a property
of the write path rather than of detection correctness, it does not depend on the
manifest partition, and re-running it unscoped over the full contract set --- 342
violations rather than 140 --- likewise created and lost no node and altered no
\prop{detectedAt}. Separately, the mediator was exercised on twenty
transition attempts, ten permitted arcs and ten violating ones; all ten
permitted arcs committed and all ten violating attempts were rejected.

\smallskip
\noindent\textbf{Co-location.} The four competency questions of
Section~\ref{sec:vsp-formal} execute as single queries against the evaluation graph and are
released with it. This is the co-location claim in operational form: Waltersdorfer
and Sabou identify the fragmentation of audit evidence across systems as the
obstacle to acting on governance obligations~\cite{waltersdorfer2025auditing}, and
none of the four is answerable from detection outputs alone. CQ1 and CQ4 ask for a
review history that stateless detection never records, CQ3 needs the lifecycle
status that separates an open finding from a closed one, and CQ2 asks about a past
instant that re-executing the rule cannot reconstruct. One caveat: CQ3's multi-entity
join returns results only for contracts, because each signatory in the evaluation
set carries exactly one violation, leaving the person arm nothing to match.

The third claim is the substantive one, because it could have failed. Keeping the
rule external and referenced only by identifier is a design decision, and whether
violations survive a revision of the rule they were recorded under is a
consequence of that decision rather than of the upsert. The rule revision widened
V3's scope from its manifest to the full contract taxonomy: the 50 original
violations retained identity and history, 135 new violations appeared only
for newly in-scope contracts, and re-running the original rule produced no
duplicates.

\subsection{Rule-Body Scope Coverage: V4 on GDPRhub Decisions}
\label{sec:eval-gdprhub}

\smallskip
\noindent\textbf{Finding.} The pattern-level claim is that audit continuity
survives a revision of the rule body. Every violation recorded under Rule A
retains its identity under Rule B through the composite key, the transition
history accumulated under Rule A remains valid under Rule B without replay or
migration, and dismissed-then-redetected traversals span both rule-body regimes.
Compliance teams need exactly this property when regulator interpretation forces
detection logic to evolve.

\begin{figure}[!htbp]
\centering
\includegraphics[width=\textwidth]{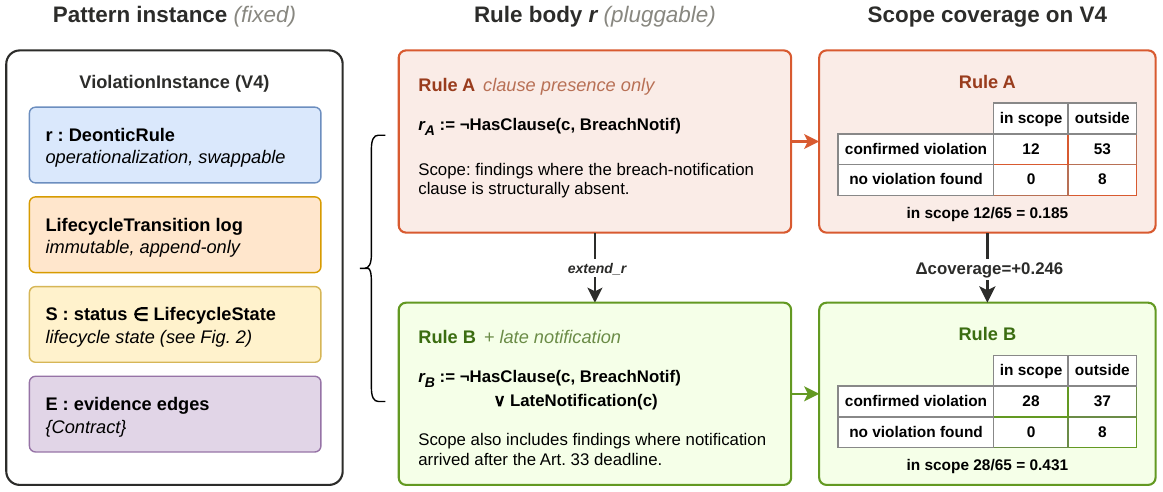}
\caption{Rule-body scope coverage on V4. The pattern instance (left,
fixed) is unchanged across two rule bodies (center, pluggable), whose scopes the
figure states. Coverage on 73 GDPRhub enforcement decisions (right) rises from
0.185 to 0.431 when $r$ is extended, with the pattern's identity, lifecycle, and
evidence semantics unchanged.}
\label{fig:ruleindep}
\end{figure}

The revision itself widens what the rule body reaches. On 73 GDPRhub Article 33
enforcement decisions, extending V4 from clause presence (Rule A) to clause
presence plus late notification (Rule B) raises the share of regulator-confirmed
findings the body encompasses from 0.185 to 0.431 (Fig.~\ref{fig:ruleindep}). Both
shares rest on subtype labels assigned by practitioners independent of the rule-body
design, but without a second annotator or an agreement measure
(Section~\ref{sec:limits}).
Both bodies are evaluated as predicates over evidence fields extracted from the
decisions --- whether a notification was made, and how long it took --- so
Rule B's deadline arm is a comparison against a threshold rather than a lookup of
a subtype label. Coverage is unchanged at thresholds of 24, 48, and 72 hours, so the figure does
not depend on the threshold chosen. This remains a scope-coverage measurement
rather than an executed detection run: the decisions are regulator case records,
not contracts in the graph.

\smallskip
\noindent\textbf{Ground-truth construction.} The synthetic V4 set is circular by
construction, so we externalize it against GDPRhub~\cite{gdprhub}: all 92
non-subcategory cases tagged Article 33 were fetched via the MediaWiki API,
and 73 yielded parseable outcomes. A transparent outcome-to-label mapping
(\textit{Violation Found}, \textit{Upheld}, \textit{Partly Upheld} to violation;
\textit{No Violation Found}, \textit{Rejected}, \textit{No further action} to
non-violation; three parser-glitch outcomes hand-resolved) gives 65 violations
and 8 non-violations. Subtype assignment combined keyword classification with
manual verification of all 65 by DiliTrust compliance practitioners against the
regulator's summary and Holding section, giving 12 missing-notification findings,
16 late-notification, and 37 other Article~33 findings that lie outside both rule
bodies by design. Of the 16 late-notification cases, 8 carry specific hour values
extracted with GPT-4.1 as a structured-extraction tool and verified by the same
practitioners against the source documents; for the other 8 the regulator confirmed
the delay exceeded 72 hours without stating a value, recorded as
\texttt{above\_threshold\_unspecified} with the hour field null.

\FloatBarrier
\subsection{Portability: SHACL Cross-Formalism Check}
\label{sec:eval-shacl}

The instantiation is a property graph because that is the stack the products run
on. The check below tests whether the same semantics survive in an RDF stack, with
both implementations operationalizing the same clause-presence condition and the
success criterion being equal firing on equal scope. We export the
\cls{Contract}+\cls{Clause} subgraph to RDF/Turtle (770 triples) and
run pyshacl 0.31 against shapes encoding V3 and V4 with
SHACL-Core~\cite{shacl2017} SPARQL targets. V1 and V2 are not encoded as SHACL
shapes: their reasoning over mandate scope and signing-date validity spans
subgraphs that a single \texttt{sh:NodeShape} cannot capture without external
SPARQL plumbing.

On identical scope the two formalizations fire on identical contract sets:
V4 agrees on all ten data-processing agreements, and V3 agrees on all fifty
contracts of the evaluation manifest. Widening V3's scope to the full contract
set raises both implementations to the same 185 contracts, again with
identical membership. The parity concerns detection alone. Persistent
identity, lifecycle state, and audit traversal are reachable in an RDF stack
only by adding custom vocabulary and orchestration on top of the standard.

\section{Limitations, Threats to Validity, and Future Work}
\label{sec:limits}
% (items 6.2 and 6.6 resolved in batch 3 D1/D2 — their subsections §7.3 and §7.4 are now filled.)
The limitations of this work fall into two groups: rule-level
boundaries and system-level properties of the property-graph
realization. We discuss each below alongside its natural extension.

\subsection{Rule-Level Limitations}
\label{sec:limits-rules}

\begin{itemize}
\item \textbf{V4 detector boundary (Sections~\ref{sec:v3v4}, \ref{sec:eval-gdprhub}).}
None of the 37 findings outside Rule B's scope is reachable by a clause-presence
or deadline predicate; reaching them requires clause-text natural-language
inference and structured Article 33(5) predicates, detector extensions that keep
the pattern unchanged.

\item \textbf{V3 normative status (Section~\ref{sec:v3v4}).} A \prop{deonticForce}
field on the rule would distinguish policy-derived from statutory obligations
without altering the pattern.
\end{itemize}

\subsection{System-Level Limitations}
\label{sec:limits-system}

\begin{itemize}
\item \textbf{Upstream extraction quality.}
Precision depends on upstream extraction (contract-signatory,
clause-boundary, clause-type), which we do not separately benchmark; a full
system evaluation would compose extraction-stage and pattern-stage error rates.

\item \textbf{Application-layer enforcement (Section~\ref{sec:vsp-graph}).}
A caller bypassing the \texttt{transition()} mediator can violate either
invariant, or both, so a deployment must confine write access to
\cls{ViolationInstance} and \cls{LifecycleTransition} to the mediator itself; the
guarantees are only as strong as that control. The natural port is an
RDF-star\,+\,SHACL realization that lifts the mediator into schema-level shapes.

\item \textbf{Operational properties.} Concurrent transitions and
production-scale ($10^5$--$10^7$ instances) traversal
performance are not evaluated. An optimistic check on \prop{lastVerifiedAt} would
serialize concurrent writes, and indexing \prop{transitionedAt} and \prop{violationId}
is the expected scaling lever for the linearly growing transition log.
\end{itemize}

\subsection{Threats to Validity}

Several threats bound what the evaluation establishes. The rule set is
author-constructed, and although the competency questions were confirmed as
representative by the practitioners who raised the underlying requirements, no
formal elicitation protocol was followed and no structured survey was conducted,
so they reflect operational query patterns rather than a systematic requirements
analysis.
Detection is manifest-scoped, so the reported detection counts characterize a
curated partition rather than the full graph; the stability results do not
inherit this bound, being reported unscoped as well.

Two of the three stability properties hold by construction, so verifying them
tests the implementation rather than the design; only the survival of identity
and history across a rule revision could have come out otherwise. The lifecycle
itself was driven by a deterministic exerciser and an invariant test, not by
reviewers working real cases, so the automaton is shown to be enforced but not
shown to fit the way review actually proceeds.

The evaluation graph is a fixture rather than client data: it instantiates the
production schema but is populated from public corpora with planted violations,
and the governance layer V1 and V2 reason over --- mandates, signing authorities,
and roughly half the signing dates --- is generated rather than observed, because
the source corpora contain none of it. V3 and V4 run over real contract text and
its clause annotations. Planting the violations is what makes ground truth
available, but it means the evaluation speaks to the pattern's behaviour and not to
detection quality on real governance data. The
instantiation is also a single domain; a schema-transfer exercise in the
artifact, re-instantiating the pattern on a GDPR-consent schema, suggests it
carries to others, but a systematic cross-domain study remains future work.

Finally, the subtype classification underlying the coverage analysis was performed
by compliance practitioners independent of the rule-body design, so the labels and
the scope they determine were not set by the same people; no measure of
inter-rater agreement was computed, however, and a second independent pass would be
needed to quantify labelling reliability.

\section{Conclusion}
\label{sec:conclusion}

This paper introduced the Violation Situation Pattern (VSP), an ontology pattern that
represents compliance violations as persistent graph entities rather than transient
query results. Each violation is modeled with a stable composite identity, a five-state
lifecycle, an append-only transition history, and rule-specific evidence links, enabling
compliance events to be tracked and audited over time. Consequently, questions such as
when a violation was first detected, who reviewed or dismissed it, and whether it has
recurred can be answered directly through graph traversal rather than by reconciling
records across multiple systems. Our evaluation demonstrates that VSP remains stable as
compliance rules evolve: repeated rule execution creates no duplicate violations, rule
revisions preserve accumulated audit history, and widening a rule's scope maintains
violation identity together with its lifecycle and evidence.

Although the current implementation relies on a mediator to preserve lifecycle
invariants during updates, future work will investigate enforcing these guarantees
directly at the schema level, for example through SHACL constraints on an RDF-star
representation. We also plan to evaluate VSP in broader governance settings beyond legal
compliance, including AI governance frameworks such as TAIR~\cite{hernandez2025tair} and
ARGOS~\cite{pathirage2025argos}, where findings similarly require persistent identities,
evidence, and lifecycle tracking. These directions will help assess the generality of
the pattern across compliance and governance domains.
\begin{acknowledgments}
All research contributions, methodological decisions, claims, and analyses are
the authors' own. The authors have no competing interests to declare that are
relevant to the content of this article.
\end{acknowledgments}

%% CEUR-WS requires a Declaration on Generative AI (in effect since January 2025).
%% See https://ceur-ws.org/genai-tax.html . The declaration below restates, in the
%% required form, the tool usage already disclosed in the acknowledgments above.
\section*{Declaration on Generative AI}
During the preparation of this work, the authors used GPT-4.1 in order to:
extract deadline values from regulator case summaries (structured extraction,
Section~\ref{sec:eval-gdprhub}); and Anthropic's Claude in order to: grammar and
spelling check, and rephrasing (editorial assistance). After using these tools,
the authors reviewed and edited the content as needed and take full responsibility
for the publication's content.

\bibliography{references}

\end{document}